# Using Feature Weights to Improve Performance of Neural Networks

**Ridwan Al Iqbal**

American International University-Bangladesh

Dhaka, Bangladesh

stopofeger@yahoo.com

**Abstract**

Different features have different relevance to a particular learning problem. Some features are less relevant; while some very important. Instead of selecting the most relevant features using feature selection, an algorithm can be given this knowledge of feature importance based on expert opinion or prior learning. Learning can be faster and more accurate if learners take feature importance into account. Correlation aided Neural Networks (CANN) is presented which is such an algorithm. CANN treats feature importance as the correlation coefficient between the target attribute and the features. CANN modifies normal feed-forward Neural Network to fit both correlation values and training data. Empirical evaluation shows that CANN is faster and more accurate than applying the two step approach of feature selection and then using normal learning algorithms.

## 1  Introduction

Feature selection is a popular method to improve performance of inductive machine learning algorithms. Many learning problems have a large feature set with many redundant features. Thus, extracting useful features for learning improves performance considerably [Guyon & Elisseeff, 2003]. However, feature selection algorithms are only preprocessors that select and modify a dataset, discarding less relevant features. After this preprocessing stage, machine learning algorithms treat all the features with equal importance. However, less-relevant features may still contribute much in the learning problem, so totally discarding them may impede accuracy; This is the reason feature selection degrades performance in some cases. While some features can be more important to the learning problem. So, a ranking based on importance towards the learning problems can be generated. In fact, such ranking measures are used in many feature selection algorithms [Leray & Gallinari, 1998; Bekkerman et al., 2003; Ruck et al., 1990].

Moreover, the principal problem in machine learning is not just having accuracy. But it is to maintain high accuracy if given less amount of data. Training data is scarce in most fields. In the data mining problems, having data is not an issue as the goal is to learn from massive data warehouses. However, most learning problems in which we have not made good progress are the ones where data is limited. Hence, high accuracy with scarce data is a must for real world use of machine learning.

But most machine learning algorithms are inductive and require large portion of data. The amount of data needed is the focus of statistical learning theory [Kearns & Vazirani, 1994]. However, developing more and more refined inductive algorithms is not the solution to the scarcity of data problem. As per learning theory, there is a fundamental limit on how much knowledge can be learned from a set of data. So, the only solution is to provide external knowledge along with data.

The focus of this paper is to learn faster if given external knowledge in the form of feature importance weight. Instead of treating all features equally, if the learners treat features based on their importance and use this knowledge of importance in learning, then they have been shown to perform better. [Iqbal, 2011; ZHANG & WANG, 2010]. This is a new field in machine learning research that has been gaining attention. *IANN (Importance aided Neural Network)* [Iqbal, 2011] extended Multilayer Perceptrons [Mitchell, 1997] to use Feature Importance values. Domain knowledge was provided as feature weights, a real value in [0,1] range to represent the importance of a feature. IANN performed better than many empirical learning algorithms. IANN also required significantly less training data to perform well. Which is much more important than improved accuracy as acquiring training data is expensive in most domains [Scott, 1991; Marcus, 1989] .

Our research uses a different approach than the IANN system. We present *CANN (Correlation aided neural network)* a neural network system that can use feature importance values in the learning process to attain better performance. It is more robust and theoretically sound than IANN. While IANN is based on heuristics with little theoretical justification, CANN is based on the same principles as Neural Network (NN) Backpropagation itself.

IANN algorithm used the feature importance values by changing the learning rate based on importance. The connections between the input features and the first hidden layer nodes had different learning rates scaled by the feature importance value. The weights were also initialized so that the more important features have higher probability of having a larger initial weight. The heuristic being, more important features have overall higher connection weight while redundant features will have less influence and thus less overall connection weight. This heuristic has been shown to be successful experimentally; even if it doesn't have theoretical foundation.

CANN uses a different design by changing the search objective instead of variation in the learning rate. CANN considers feature weights to be the correlation between input features and the output. The algorithm thus tries to fit the given correlation as well as fitting data. Thus, CANN is adding additional constraints into Backpropagation.

In this paper, CANN is applied in several real world datasets of different domains. The performance is analyzed and compared with many popular learning algorithms. Brief outline of the paper: Section 2 provides the theoretical background behind CANN, section 3 describes the actual algorithm while section 4 shows the experimental results. Section 5 concludes the paper.

We now describe the notation used in this paper. N is the number of instances while K is the number of features. $I_k$, is used to refer the feature importance of feature $X_k$. $X$ represents the set of all features in the data set $D$ of a learning model $L$. cov(x,y) is the covariance between x and y. The output from a neural network is $y(X)$ given an instance x. Consequently, y also represents the actual output node whenever mentioned in a subscript. The components of neural networks are represented in a layered fashion. A superscript always represents the source layer while subscripts represent nodes between successive layers. Therefore, $w_{ij}^k$ is the weight from node i of layer k to node j of layer k+1. In the fashion, $h_j^n$ is the output of a node j in layer n. $\theta_j$ is the bias of node j. $\delta_j$ represents the error of unit j; $\alpha$ is the learning rate. *Layer$_k$* or $L_k$ is the set of units of layer k, where $k$ is 1 for the first hidden layer; *Inputs(k)* is the set of all units that is input to $k$. $\mu$ represents mean. $U$ is the set of all network units.

## 2    Description of CANN

The notion that some features are more important to a learning problem than others is central to machine learning. There has been many research that attempts to provide a measure of importance e.g FIRM [Zien et al., 2009]; Saliency [Ruck et al., 1990]. The goal of these researches primarily was to aid feature selection. However, instead of calculating importance, our algorithm is given the importance values. And we would like to quantify that importance into something in the learning model. Therefore, before we can use the feature importance values we must choose what feature importance means to a dataset.

Correlation coefficient is a very popular measure of the relationship between two random variables [Rodgers & Nicewander, 1984]. Correlation coefficient measures how two variables linearly influence each other. In other words, if both variables are strongly correlated, then they are expected to change together. The correlation coefficient between a feature and the target variable will measure how much influence a feature has over the target variable. This is a very good measure of importance for a feature.

We consider feature importance to be the correlation coefficient between the output variable and the input feature in consideration.

**Definition 1.** *Feature Importance $I_k$ for a feature $X_k$ is the true correlation coefficient between output variable y and $X_k$.*

$$I_k = corr(y, X_k) = \frac{cov(y, X_k)}{\sigma_y \sigma_{X_k}} \tag{1}$$

$$corr(y, X_k) = \frac{cov(y, X_k)}{\sigma_y \sigma_{X_k}} \tag{2}$$

$$= \frac{E[(X_k - \mu_{X_k})(y - \mu_y)]}{\sigma_y \sigma_{X_k}}$$

$$cov(y, X_k) = E[(X_k - \mu_{X_k})(y - \mu_y)]$$

$$= E(X_k y) - E(y) E(X_k)$$

True mean can be estimated based on the dataset as sample mean:

$$cov(y, X_k) = \frac{1}{n} \sum_i^N (X_k^i - \overline{X_k})(y^i - \bar{y})$$

$$= \overline{X_k y} - \overline{X_k} \cdot \bar{y} \tag{3}$$

CANN algorithm is given both the training data and the correlation values of features. The feature importance values can be derived in different ways. It can be deduced by experts or it can be calculated from other different datasets. Normal Multilayer Perceptrons (MLP) only fits the training dataset. Our goal in CANN algorithm is that the learned neural network fits both the dataset and the correlation values given. Thus, we would like to minimize correlation error as well.

Let *L(I, D)* be the learned model over training set D and Feature Importance set I. The learned model L will have an output for each instance in the dataset from which correlation coefficient can be generated. Feature Importance is the target correlation value we want the learned model to have over the training set.

We define correlation error $E_c$ to be the squared error between target correlation value $I_k$ and the correlation coefficient between the output of the learned model y and input set X over training set D.

$$E_c(D,I) = \frac{1}{2}\sum_k [I_k - corr(y, X_k)]^2 \qquad (4)$$

However, dataset for training is already given before learning begins. Thus $\sigma_y \sigma_{X_k}$ is a constant for this particular training set and can be calculated beforehand from the data. As $I_k$ is considered the target correlation value, so using (1) we can see that instead of fitting $I_k$ we can fit.

$$c_k = \sigma_y \sigma_{X_k} I_k$$

$$E_c(D,I) = \frac{1}{2}\sum_k [c_k - cov(y, X_k)]^2 \qquad (5)$$

Multilayer Perceptrons or feed-forward neural networks have neurons or nodes that connected in a feed forward fashion in many layers. Connection is allowed only between successive layers. It normally has 1 input and output layer and N hidden layers. The logistic activation function $\sigma(x) = (1 + e^{-x})^{-1}$ is the most popular. The nodes are then trained to minimize error in a backward manner. This algorithm is known as Backpropagation which tries to minimize least squared error over dataset using stochastic gradient descent to find the suitable weights for the connections between nodes [Mitchell, 1997].

The general form of a feed-forward neural network is:

$$y(\mathcal{X}) = \sigma\left[\sum_{j \in L_n} w_{jy}^n h_j^n + \theta_y\right] \qquad (6)$$

$$h_j^k = \sigma\left[\sum_{i \in L_{k-1}} w_{ij}^{k-1} h_i^{k-1} + \theta_j\right]$$

$$h_j^0 = X_j$$

The training rule for general Backpropagation is derived by taking a partial derivative of the training error $E_D$.

$$\Delta E_D = \frac{1}{2}\frac{\partial E_D}{\partial w_{jy}^n} = \frac{\partial}{\partial w_{jy}^n}\sum_{d \in D}(t_d - y_d)^2 \qquad \forall w_{jy}^n \in W^N$$

$$= -\sum_{d \in D}(t_d - y_d)y_d(1-y_d)h_j^N$$

Here, $\frac{\partial}{\partial w_{jy}^n} y_d = y_d(1-y_d)h_j^N$

$t_d$ = target value of the input d;
$y_d$ = output of the network for input d;
$W^N$ = Set of weights connected with the output layer.

Our algorithm works by changing the error function of MLPs to minimize correlation error as well. We extend the error function of normal Backpropagation to include $E_c$ as well. Therefore, instead of just minimizing data set error, Backpropagation would minimize both. We would also add a scaling constant $p \to [0,1]$ to control the weight given to correlation based error $E_c$. Thus,

$$E = pE_D + (1-p)E_c$$

$$\Delta E = p\frac{\partial E_D}{\partial w} + (1-p)\frac{\partial E_C}{\partial w} \qquad \forall w \in W^N$$

$$\frac{\partial E_C}{\partial w_{jy}^n} = \sum_k (c_k - cov(y, X_k))\left(-\frac{\partial}{\partial w}cov(y, X_k)\right) \qquad (7)$$

The derivative of the covariance can be derived to be:

$$\frac{\partial}{\partial w_{jy}^n} cov(y, X_k) = \frac{\partial}{\partial w_{jy}^n}\frac{1}{n}\sum_i^N X_k Y - \overline{X_k}\frac{1}{n}\sum_i^N Y$$

$$= y_d(1-y_d)\left[\frac{1}{n}\sum_d x_k h_j^N - \overline{X}.\frac{1}{n}\sum_i^N h_j^N\right]$$

$$= y_d(1-y_d)[\overline{x_k . h_j^N} - \overline{X}.\overline{h_j^N}] \qquad (8)$$

$$\frac{\partial E_C}{\partial w} = -\sum_k (c_k - cov(y, X_k))y_d(1-y_d)[\overline{x_k . h_j^N} - \overline{X}.\overline{h_j^N}] \qquad (9)$$

The training rule for normal Backpropagation is normally derived using stochastic gradient descent so that the whole dataset is not iterated to make one change [Mitchell, 1997]. The training for output layer calculates the error which is then propagated backward into lower layers. The error for output in simple Backpropagation is simply a changed $-E_D$ which is not summed over all dataset,

$$\delta_d^* = (t_d - y_d)y_d(1-y_d)h_j^N$$

Hence, for CANN we can also use stochastic gradient descent. Thus the error term for CANN:

$$\delta_d = y_d(1-y_d)\left\{\begin{array}{l} p(t_d - y_d)h_j^N + \\ (1-p)\sum_k (c_k - cov(y, X_k))(\overline{x_k . h_j^N} - \overline{X_k}.\overline{h_j^N}) \end{array}\right\}$$

Therefore, this is the new error function used by Backpropagation.

$$\Delta w_{ij}^k = \begin{cases} \alpha \delta_d; & \text{for Layer}_N \\ \alpha X_{ij}^k \sum_{p \in L_{k+1}} \delta_p^{k+1} w_{jp}^{k+1}; & \text{for other layers} \end{cases} \qquad (10)$$

## 3  Implementation of CANN

The training rule of CANN includes sample Covariance and sample Mean terms that must be calculated over the entire dataset each time a single instance is iterated. This is computationally very costly. We cannot pre-calculate the Mean terms because they depend on the output of the network for each instance; which is in turn dependent on the configuration of the network. Each time a weight is updated, the sample Mean also changes.

So, instead of calculating the means every time, we use Memoization and iterative improvement to calculate Mean. This requires some space, but the order of the space complexity remains unchanged.

The idea is to keep a table or array that stores the element of the mean indexed by instance number. Sample Mean is nothing but the weighted sum of all the element points. So,

$$\bar{A} = \frac{A_1}{n} + \frac{A_2}{n} \ldots \frac{A_n}{n}$$

Thus, if the output value of one training instance is changed, the changed Mean can be calculated just by subtracting the previous value and adding the new value. This is the reason for storing element values in a table. For each instance in the training set, we calculate the new value of the network and then replace the value in the table with the new value and also change the Mean by subtract-add formula just described.

For example, we require $\bar{Y}$ to calculate covariance according to (3). The table $T_y$ holds weighted output of the network for each training instance.

$$T_y^i := \frac{y(i)}{n} \qquad \forall i \in D$$

Initially, after initializing the weights of the network we calculate the initial output for each data point, populate $T_y$ and calculate the initial Mean $\bar{y}$. During Backpropagation training, we will calculate new Mean as:

$$\bar{y} := \bar{y}^* - T_y^i + \frac{y(i)}{n}$$

This moving average is not the real Mean, but it will eventually reach convergence after several training epochs through the dataset. This is similar to the stochastic gradient descent approach used for updating weights.

We will calculate $\overline{X_k Y}$, $\overline{x_k . h_j^N}$, $\overline{h_j^N}$ & $\bar{y}$ in this way. The space required for each mean calculation is O(N) where N is the number of instances in the training set. If there are K features then overall space required will be O(NK) which is also the space complexity for simple MLP.

## 4 Empirical evaluation

We tested CANN with datasets from the University of California, Irvine (UCI) repository. The datasets chosen have relatively higher number of features and are also known to be especially difficult. The data sets used are described below:

1. **Soybeen-Large:** Instances: 683, Attributes: 35, All nominal, 19 classes.
2. **Spambase:** Instances: 4601 (1813 Spam = 39.4%), Attributes: 58 (57 continuous, 1 nominal class label).
3. **Promoter Gene Sequences:** Instances: 936(236 positive), Attributes: 58, nominal, 2 classes and Expert Domain theory.
4. **Cardiac Arrhythmia:** Instances: 452, Attributes: 279, mostly real including some binary, 16 classes.
5. **Annealing:** Instances: 798, Attributes: 38, 29 nominal and the rest numeric, 6 classes.

The algorithms compared include both classic and new but high performing ones. We tested standard Feed-forward Neural Network [Mitchell, 1997], C4.5 [Quinlan, 1993], Support Vector Machines [Vapnik, 1998], K-Nearest Neighbour, and Naïve Bayes [Mitchell, 1997a].

| Methods | Soybeen | Spambase | Promoter | Arrhythmia. | Annealing |
|---|---|---|---|---|---|
| IANN | 87.45 | 89.57 | 87.83 | 55.73 | 90.32 |
| MLP | 86.80 | 89.76 | 85.33 | 62.44 | 88.23 |
| **CANN** | **89.05** | **91.35** | **91.42** | **69.16** | **91.39** |
| C4.5 | 82.69 | 91.52 | 81.98 | 66.81 | 89.30 |
| SVM | 87.35 | 80.91 | 85.23 | 52.76 | 81.95 |
| Near. N. | 85.04 | 88.78 | 85.64 | 57.97 | 88.86 |
| Nai. Bay | 88.32 | 78.22 | 94.30 | 64.15 | 73.71 |
| Best result | CANN | C4.5 | Nai.Ba | CANN | CANN |

*Table 1: Experiment results in percent Accuracy (50% test-set)*

In order to have a valid correlation based knowledge, we decided to calculate the correlation of features using the full dataset. This was done for all the datasets except promoters which has an associated Expert knowledge. The Importance values for Promoters dataset was derived from that knowledge. These correlation values were given to CANN. The algorithms were then trained using 50% of the data and the rest was used as test set. Experiments are averaged over 20 iterations. The results are shown in Table 1.

| Methods | Soybeen | Spambase | Promoter | Arrhythmia. | Annealing |
|---|---|---|---|---|---|
| IANN | 87.45 | 89.57 | 87.83 | 55.73 | 90.32 |
| MLP | 87.56 | 91.76 | 86.75 | 62.38 | 91.47 |
| **CANN** | **89.05** | **91.35** | **91.42** | **69.16** | **91.39** |
| C4.5 | 80.93 | 90.78 | 85.04 | 64.60 | 87.75 |
| SVM | 89.35 | 81.91 | 86.75 | 50.88 | 84.85 |
| Near. N. | 83.51 | 90.65 | 85.64 | 48.23 | 90.97 |
| Nai. Bay | 88.85 | 78.65 | 94.53 | 59.73 | 77.95 |
| Best result | CANN | MLP | Na.Bay | CANN | CANN |

*Table 2: Experiment result: 85% feature selection (50% train set)*

The results show clear difference between CANN and normal Neural Network. CANN outperforms MLP significantly in all datasets. The difference is more pronounced in the difficult datasets such as Arrhythmia or Promoters where the average performance over all the algorithms is lower. CANN generally outperforms all other algorithms. Except in Promoters and Spambase where a particular algorithm outmatches all others.

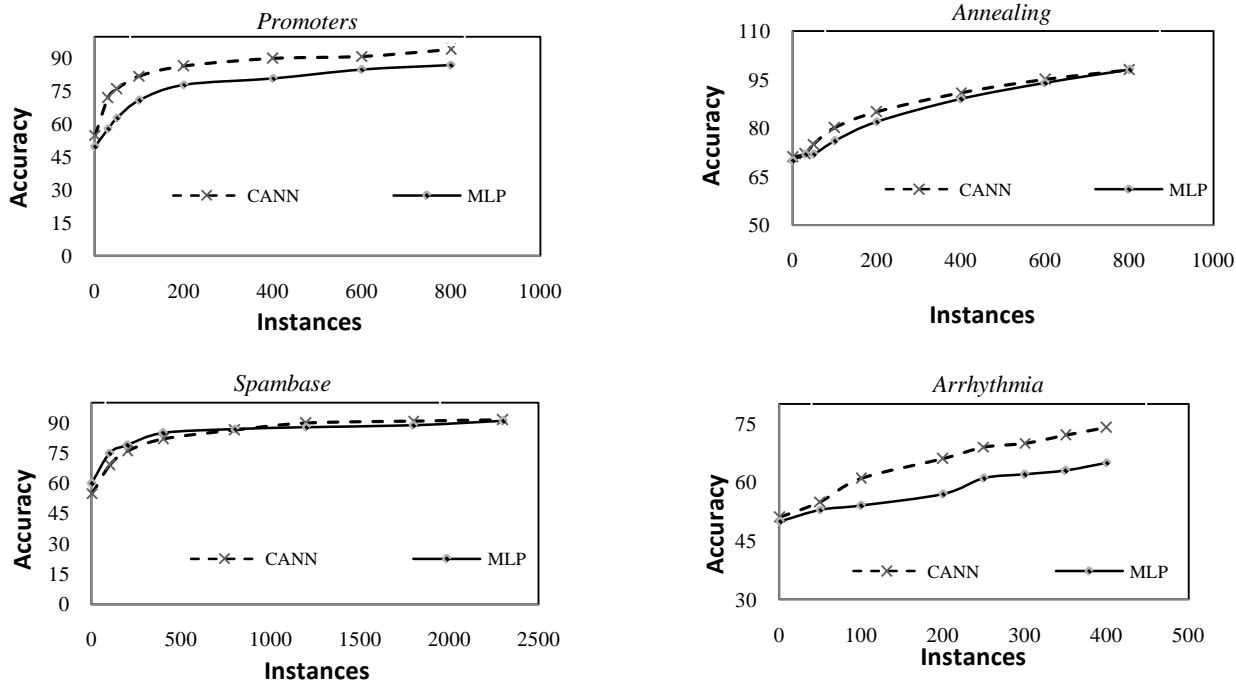

*Figure 1: The learning curve for 4 datasets*

This reason is a problem can be specifically suitable for a particular algorithm. But even then CANN performs next best. The performance difference between and IANN and MLP is not statistically significant. IANN outperforms MLP in some datasets slightly. However, no general trend emerges. So, CANN clearly makes a better use the feature importance weight given.

The next experiment shown in Table 2 is when feature selection was used for the inductive algorithms while CANN and IANN used the full feature set. 15% features were removed using Chi-squared evaluation ranking. 50% of the datasets was used for training and the rest for tests.
The results in Table 2 show that using feature selection does not always improve performance. Apart from few changes the performance makeup almost remained unchanged. Some of the algorithms had a better performance in a dataset while the performance actually decreased in some. There is no uniform trend. The results for MLP slightly increased in some cases. It surpassed CANN only in the Spambase dataset. But overall the advantage of CANN over normal MLP and other algorithms remained. CANN performed the best on average.

Another insight is that IANN was outperformed by MLP when feature selection was used. So, feature selection and MLP combination was better than IANN But not better than our algorithm.

In the figure 1, the learning curves are shown for the datasets. The comparison is between MLP and CANN. It is evident from the curves that CANN initially performs better and learns faster. As dataset size increases, MLP eventually closes the performance gap. However, this trend was not true in all cases, the performance gap depended on the difference between the correlation value provided to CANN and the correlation makeup of the dataset. If the correlations of the dataset are close to true correlation then CANN will show less performance improvement as there will be no knowledge advantage; which is the case for Spambase which performs better than CANN. On the other hand, The feature weights for promoters were derived from an expert domain theory. So, it always outperforms normal MLP.

## 5   Related works

The research on using different error function other squared error has been comprehensive. Many different and seemingly exotic error functions have been tried with success [Haykin, 1998]. However, the use of domain knowledge instead of training on data only has been limited. The KBANN family of algorithms incorporated rule based domain theory by initializing both the network structure and the weights [Towell & Shavlik, 1994]. The use of additional constraints along with the tradition error function has been explored before. This is called constraint based learning. Certain derivatives of the target function can be specified in prior. This approach has been explored by Simard et al. in TangentProp [Simard et al., 1992] which provided additional constraint to Backpropagation to fit the input derivatives. Knowledge-based SVM used domain knowledge provided as additional constraints into Support Vector Machines [Fung et al., 2002]. Abu Mustafa showed how to provide hints as additional constraint [AbuMostafa, 1995].

## 6  Conclusion

We have proposed a well performed approach of incorporating feature importance into neural network learning. The performance of such a learner shows feature importance aided learners can achieve superior performance over ordinary inductive learners. Removing irrelevant features by feature selection is a good approach, however Expert knowledge is available in some domains or correlation of same features could be calculated from a different problem dataset as well. This extra knowledge could be transferred to CANN to attain higher performance. This approach of incorporating feature importance into learners is worthy of further development. Possible future applications of this algorithm will be areas where related machine learning problems are being solved or where expert knowledge is available. The future research areas can be modifications of existing popular empirical learners so that they utilize feature importance. Correlation coefficient aided algorithms maybe developed for algorithms such as Support Vector Machines; Decision Tree based algorithms or Bayesian classifiers.

Current machine learning algorithms rely too much on training examples. Incorporating more and more domain knowledge or using the knowledge from related problem area is the way for improvement. Our proposed method shows how improvements can be had from such methods.